\PassOptionsToPackage{numbers, compress}{natbib}

\documentclass{article}

\usepackage[preprint]{neurips_2023}

\usepackage[utf8]{inputenc} 
\usepackage[T1]{fontenc}    
\usepackage{hyperref}       
\usepackage{url}            
\usepackage{booktabs}       
\usepackage{amsfonts}       
\usepackage{nicefrac}       
\usepackage{microtype}      

\usepackage{array}
\usepackage{tabularx}
\usepackage{tcolorbox}
\usepackage{graphicx}
\usepackage{caption}
\usepackage{subcaption}
\usepackage{bm}
\usepackage{amsmath,amssymb}
\usepackage{booktabs}
\usepackage{algorithm}
\usepackage[noend]{algorithmic}
\usepackage{multirow}
\usepackage[flushleft]{threeparttable}

\usepackage{multirow}
\usepackage{makecell}
\usepackage{multirow}
\usepackage{xspace}
\usepackage{color}
\usepackage{wrapfig}
\usepackage{bbding}

\usepackage{hyperref}
\usepackage{url}
\usepackage{amsthm}
\usepackage{amsmath}

\usepackage{pifont}

\usepackage{xcolor}
\definecolor{themecolor}{RGB}{214, 226, 239}

\usepackage{tcolorbox}
\usepackage[frozencache,cachedir=.]{minted}
\tcbuselibrary{minted, breakable, xparse, skins}

\definecolor{bg}{gray}{0.95}
\DeclareTCBListing{mintedbox}{O{}m!O{}}{%
  breakable=true,
  listing engine=minted,
  listing only,
  minted language=#2,
  minted style=friendly,
  minted options={%
    linenos,
    gobble=0,
    breaklines=true,
    breakafter=,,
    fontsize=\small,
    numbersep=8pt,
    #1},
  boxsep=0pt,
  left skip=0pt,
  right skip=0pt,
  left=25pt,
  right=0pt,
  top=3pt,
  bottom=3pt,
  arc=5pt,
  leftrule=0pt,
  rightrule=0pt,
  bottomrule=2pt,
  toprule=2pt,
  colback=bg,
  colframe=themecolor,
  enhanced,
  overlay={%
    \begin{tcbclipinterior}
    \fill[themecolor] (frame.south west) rectangle ([xshift=20pt]frame.north west);
    \end{tcbclipinterior}},
  #3}

\title{TAGLAS: An atlas of text-attributed graph datasets in the era of large graph and language models}

\author{
    Jiarui Feng$^{1}$~~~Hao Liu$^{1*}$~~~Lecheng Kong$^{1}$\thanks{Contributed equally. Listing order is random.}~~~Mingfang Zhu$^{2}$~~~Yixin Chen$^{1}$~~~Muhan Zhang$^{3}$\thanks{Corresponding author}\\
    \texttt{\{feng.jiarui, liuhao, jerry.kong, ychen25\}@wustl.edu, lisazhu@NMSU.edu, }\\
    \texttt{muhan@pku.edu.cn} \\
    ${}^1$Washington University in St. Louis~~~${}^2$New Mexico State University~~~${}^3$Peking University\\}

\begin{document}

\maketitle

\begin{abstract}
In this report, we present TAGLAS, an atlas of text-attributed graph (TAG) datasets and benchmarks. TAGs are graphs with node and edge features represented in text, which have recently gained wide applicability in training graph-language or graph foundation models. In TAGLAS, we collect and integrate more than 23 TAG datasets with domains ranging from citation graphs to molecule graphs and tasks from node classification to graph question-answering. Unlike previous graph datasets and benchmarks, all datasets in TAGLAS have a unified node and edge text feature format, which allows a graph model to be simultaneously trained and evaluated on multiple datasets from various domains. Further, we provide a standardized, efficient, and simplified way to load all datasets and tasks. We also provide useful utils like text-to-embedding conversion, and graph-to-text conversion, which can facilitate different evaluation scenarios. Finally, we also provide standard and easy-to-use evaluation utils. The project is open-sourced at \url{https://github.com/JiaruiFeng/TAGLAS} and is still under construction. Please expect more datasets/features in the future.
\end{abstract}

\section{Introduction}
In recent years, research on graph learning has been growing at a fast speed. We have witnessed the emergence of many advanced graph models, applications, and benchmarks. Particularly, graph neural networks (GNNs)
have become one of the dominant methods for graph learning with the ability to handle node-level~\citep{GCN, SGC, graphsage, GIN}, link-level~\citep{SEAL, neognn,nbfnet, buddy, ncnc}, and graph-level tasks~\citep{dgcnn, ngnn, kpgnn, gdgnn, de, kwlgnn, maggnn, n2gnn} on graphs. Along with the development of the GNN models, many benchmarking datasets are proposed to facilitate the research in this field. For example, citation networks Cora, Pubmed, and Citeseer~\citep{cora} are commonly used benchmarking datasets for node classification. Knowledge graphs like FB15K237 and WN18RR~\citep{fb15k} can evaluate the performance of GNNs on relational data. TU dataset~\citep{tu} collects multiple graph sets for facilitating research on graph-level tasks. More recently, Open Graph Benchmark (OGB)~\citep{OGB} provided one of the first large graph benchmark collection. OGB contains more than 10 different graph datasets, including citation networks, molecular graphs, protein-protein interactions, and knowledge graphs. They provide a unified protocol to access, run, and evaluate different graph models on these datasets. 

Beyond machine learning on graphs, the research in the language and vision fields is undergoing a huge transformation. In particular, the emergence of the concept of the foundation model turned the whole community from developing small-scale models for specialized tasks to training large-scale, versatile foundation models~\citep{bommasani2021opportunities}. For example, well-trained large language models have shown superior performance on various downstream tasks on language, even without the need for fine-tuning~\citep{gpt3, instructGPT, llama}. 

However, all existing datasets and benchmarks for graphs are extremely diverse and discrepant. Firstly, different datasets come from different domains, spanning from citation networks to 2D molecules. Datasets from different domains often contain different feature spaces. For instance, node features in citation networks are often titles and abstracts of papers. Instead, node features in molecule graphs might be one-hot encodings of chemical atoms and bonds. In addition, even for datasets from the same domain, features can be represented in different ways. For example, the node features in the Cora dataset~\citep{cora} are bag-of-words representations, while ogbn-arxiv~\citep{OGB} uses sentence embeddings generated from a language model. Secondly, different datasets focus on different tasks, which requires previous graph models to train a different classification/regression head for every task. Because of these discrepancies, the current research on graph machine learning is mostly specialized, small-scale, and hard to transfer. In the context of developing large-scale foundation models instead of specialized models, there is an urgent need for a large-scale and unified-format graph dataset collection for both training and evaluation. 

In this report, we present TAGLAS, a new graph dataset atlas that is suitable for tackling the aforementioned challenges. Specifically, in TAGLAS, we collect more than 23 graph datasets ranging from different domains and different task types. Particularly, inspired by the recent progress in graph foundation models~\citep{OFA, GraphGPT, zhang2024graphtranslator}, we unify all datasets from different domains by representing them as Text-Attributed Graphs (TAGs). In TAGs, the features of all nodes and edges are represented by plain text, which enables a single model to be trained simultaneously across graphs from different domains~\citep{OFA}. Further, to support fast training and evaluation, we provide a one-line solution for task generation. Particularly, we implement the generation pipelines for several commonly adopted task formats like rooted-subgraph extraction. We also provide many useful APIs like text-to-embedding and graph-to-text conversions, which can be used to support different training and evaluation scenarios. Finally, we also provide standard evaluation tools for each dataset. We hope the presence of TAGLAS can contribute to the research on graph-language multi-modal models and graph foundation models. 
\section{Overview}
\begin{figure}[t]
    \centering
    \includegraphics[width=1.0\textwidth]{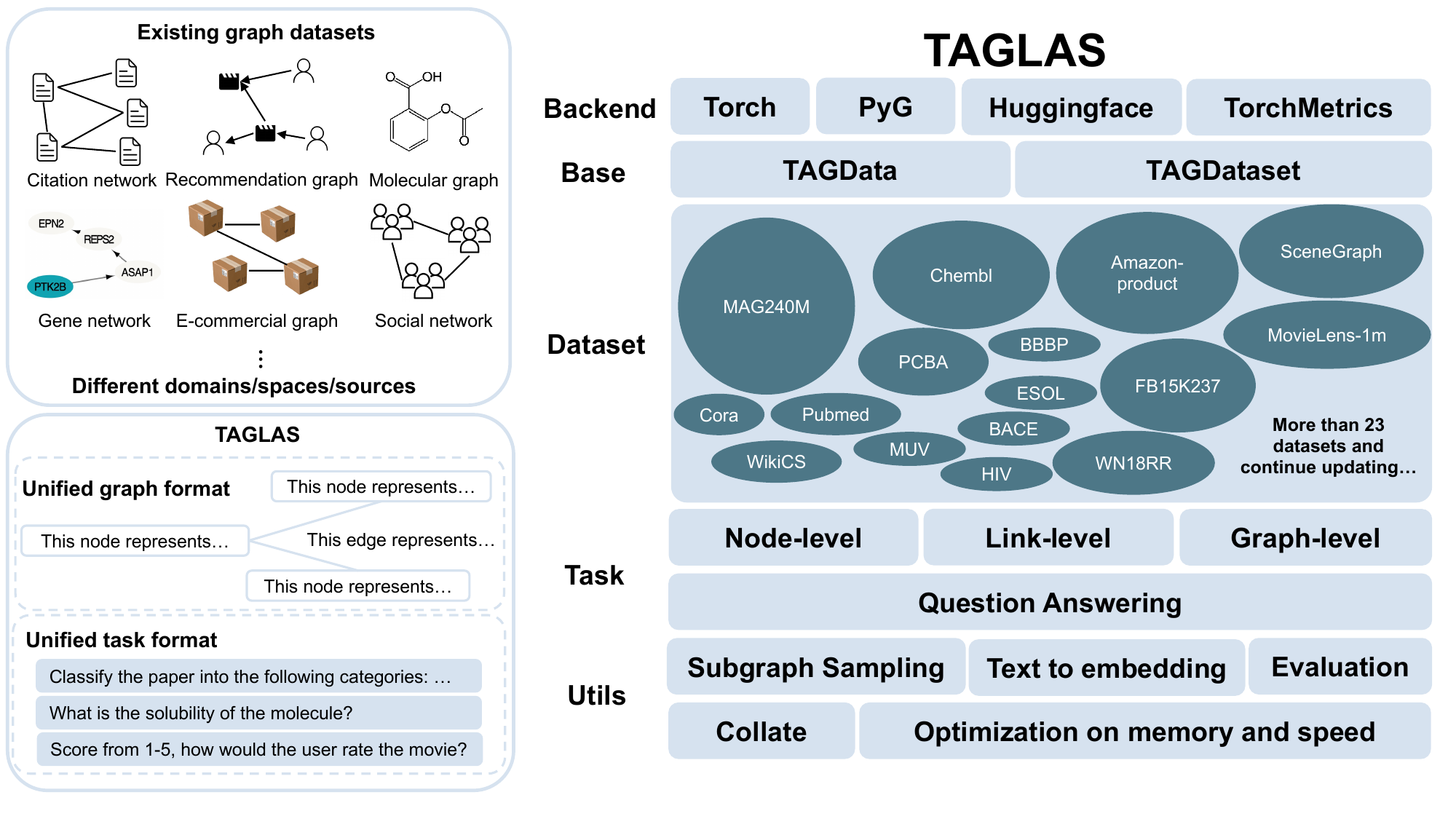}
    \caption{Overview of the TAGLAS.}
    \label{fig:overview}
\end{figure}

\begin{table}[h]
\small
 \setlength{\tabcolsep}{2.3pt}
 \renewcommand{\arraystretch}{1}
    \caption{Dataset statistics of TAGLAS. (W. represent word.)}
    \vspace{5pt}
    \label{tab:dataset}
    \centering
    \begin{tabular}{l|cccccccc}
        \toprule
        Dataset &Avg. \#N &Avg. \#E &Avg. \#N. W. &Avg. \#E. W.& \# G & Domain & Source\\ \midrule
        Cora & 2,708 & 21,112 & 143.4 & 8.0 & 1 & Co-citation & \citep{OFA}~\citep{Graphllm}\\
        Pubmed & 19,717 & 44,338 & 246.2 & 8.0 & 1 & Co-citation & \citep{OFA}~\citep{Graphllm}\\
        Arxiv & 169,343 & 1,166,243 & 174.7 & 7.0 & 1 & Citation & \citep{OGB}~\citep{OFA}\\
        WikiCS & 11,701 & 216,123 & 421.2 & 6.0 & 1 & Wikipedia page & \citep{wikics}~\citep{OFA}\\
        Product-subset & 54,025 & 144,638 & 113.9 & 6.0 & 1 & Co-purchase & \citep{TAPE} \\
        FB15K237 & 14,541 & 310,116 & 20.1 & 8.4 & 1 & Knowledge graph & \citep{OFA} \\
        WN18RR & 40,943 & 93,003 & 23.3 &11.0 & 1 & Knowledge graph & \citep{OFA} \\
        MovieLens-1m & 9,923 & 2,000,418 & 14.8 & 9.5 & 1 & Moive rating & \citep{pyg} \\
        Chemblpre & 25.87 & 55.92 & 44.0 & 15.0 & 365,065 & Molecular & \citep{gimlet} \\
        molproperties &  25.57  &  55.32  & 48.0 & 19.0 & 363,336 & Molecular & \citep{gimlet} \\
        PCBA & 25.97 & 56.20 & 48.1 & 19.0 & 437,929 & Molecular & \citep{gimlet} \\
        HIV & 25.51 & 54.94 & 44.0 & 15.0 & 41,127 & Molecular & \citep{gimlet} \\
        BBBP & 24.06 & 51.91 & 48.1 & 19.0 & 2,039 & Molecular & \citep{gimlet} \\
        BACE & 34.09 & 73.72 & 48.1 & 19.0 & 1,513 & Molecular & \citep{gimlet} \\
        toxcast & 18.76 & 38.50 & 44.1 & 15.0 & 8,575 & Molecular & \citep{gimlet} \\
        esol & 13.29 & 27.35 & 44.0 & 15.0 & 1,128 & Molecular & \citep{gimlet} \\
        freesolv & 8.72 & 16.76 & 48.1 & 19.0 & 642 & Molecular & \citep{gimlet} \\
        lipo & 27.04 & 59.00 & 48.1 & 19.0 & 4,200 & Molecular & \citep{gimlet} \\
        cyp450 & 24.52 & 53.02 & 48.1 & 19.0 & 16,896 & Molecular & \citep{gimlet} \\
        tox21 & 18.57 & 38.59 & 48.1 & 19.0 & 7,831 & Molecular & \citep{gimlet} \\
        muv & 24.23 & 52.56 & 48.0 & 19.0 & 93,087 & Molecular & \citep{gimlet} \\
        ExplaGraphs & 5.17 & 4.25 & 5.1 & 5.3 & 2,766 & Common sense & \citep{gretriever} \\
        SceneGraphs & 19.13 & 68.44 & 20.1 & 9.8 & 100,000 & scene graph & \citep{gretriever} \\
        MAG240M-subset & 5,875,010 & 26,434,726 & 152.3 & 11.0 & 1 & Citation & \citep{OGB-lsc} \\
        Ultrachat200k & 3.72 & 2.72 & 143.9 & 9.5 & 449,929 & Conversation & \citep{ultrachat} \\
        Wikikg90m & 91,230,610 & 1,202,155,622 & 18.99 & 25.70 & 1 & Knowledge graph & \citep{OGB-lsc} \\
        \bottomrule
    \end{tabular}
\vspace{-18pt}
\end{table}

In this section, we briefly describe the design of the TAGLAS. Figure~\ref{fig:overview} shows the overall structure. The TAGLAS is implemented based on Pytorch~\citep{pytorch}, PyG~\citep{pyg}, Huggingface~\citep{hf_datasets, hf_transformer}, and torchmetrics~\citep{torchmetrics}. Upon it, each TAG graph is represented by \textit{TAGData}, and each dataset is represented by \textit{TAGDataset}. Next, we collect more than 23 datasets from different domains, ranging from citation networks to molecule graphs. Each dataset can be used for one or multiple tasks. There are four task types, including the traditional \textit{node-level, link-level, graph-level} tasks, and one additional \textit{question answering} task for handling arbitrary free-form graph tasks expressible by text. In TAGLAS, some datasets are naturally designed for question-answering tasks. For other datasets, we also provide a unified protocol to convert the original classification/regression tasks to question-answering format. Further, TAGLAS supports two major task formats for node/link-level tasks: the default format and the subgraph format. The default format directly returns the whole graph for training and evaluation, which is commonly adopted in the graph community. Recently, the subgraph-based format has become popular due to its unified task representation and improved expressivity~\citep{allinone, liu2023graphprompt, OFA, zhang2021labeling}. Particularly, for each target node/link, the subgraph task format samples a rooted subgraph (ego-network) from the target node/link and uses the sampled subgraph as the input to the model. Finally, for each dataset, we provide a default API for evaluating the performance of the model. We implemented two evaluation modes to support both the tensor output and text output. In the following sections, we will describe each component of TAGLAS in detail.  
\section{Dataset collection and processing}
In this section, we describe the collection and pre-processing procedure of each included dataset. Table~\ref{tab:dataset} presents the overall statistics of each dataset.
\subsection{Cora}
The Cora dataset is a co-citation graph of computer science papers. The dataset is collected from OFA~\citep{OFA} and the original source is from Graph-LLM~\citep{Graphllm}. In Graph-LLM, authors re-collect the dataset, as the commonly used Cora dataset in the GNN community~\citep{GCN} uses bag-of-words features and the raw text is hard to retrieve. The new Cora contains 2,708 nodes and 10,556 edges, the same as the original dataset. The text feature format of nodes and edges is shown in the below:
\begin{tcolorbox}[boxsep=0mm,left=2.5mm,right=2.5mm, colback=themecolor]
\textbf{Node text:} Academic paper with title and abstract: $<$\textit{title}$>$$<$\textit{abstract}$>$.\\
\textbf{Edge text:} Connected papers are cited together by other papers.
\end{tcolorbox}
Cora dataset can be used for node classification and link prediction. For node classification, the task is to predict the category of the paper. There are 7 categories: Theory; Reinforcement learning; Genetic algorithms; Neural networks; Probabilistic methods; Case-based; and Rule learning. The train/val/test split is 140/500/2068. Note that the split differs from the split of the original dataset~\citep{GCN}, as the mapping from the original one to the new one is hard to collect. For link prediction, we follow OFA~\citep{OFA} and randomly split all edges into train/val/test with a ratio of 0.85/0.05/0.10. Meanwhile, we sample the same number of negative edges for each set to construct the negative sets. This results in a total of 17,944/1,056/2,112 samples for train/val/test sets, respectively.

\subsection{PubMed}
PubMed dataset is a co-citation graph of biomedical papers focused on diabetes mellitus. The source and the process procedure of Pubmed are the same as Cora's. After the processing, there are 19,717 nodes and 88,648 edges. Similarly, the Pubmed data can be used for node classification and link prediction. For node classification, there are 3 different categories, including Diabetes mellitus, experimental; Diabetes mellitus, type 1; and Diabetes mellitus, type 2. The train/val/test split is 60/500/19,157. For link prediction, the split procedure is the same as Cora, which results in a total of 150,700/8,866/17,730 samples for train/val/test set, respectively.

\subsection{Arxiv}
Arxiv dataset is a citation graph of papers from the arXiv platform. We collect the Arxiv dataset and its raw text from OGB~\citep{OGB} and OFA~\citep{OFA}. There are 169,343 nodes and 1,166,243 edges in the graph.  The text feature format of nodes and edges is shown in the below:
\begin{tcolorbox}[boxsep=0mm,left=2.5mm,right=2.5mm, colback=themecolor]
\textbf{Node text:} Academic paper. Title: $<$\textit{title}$>$. Abstract: $<$\textit{abstract}$>$.

\textbf{Edge text:} The connected two papers have a citation relationship.
\end{tcolorbox}
Arxiv dataset can be used for node classification. There are 40 categories for Arxiv. We directly obtain the split from OGB~\citep{OGB}, and there are 90,941/29,799/48,603 samples for the train/val/test sets, respectively.

\subsection{WikiCS} 

WikiCS is a graph generated from the Wikipedia platform. Nodes in WikiCS are Wikipedia page descriptions and edges are hyperlinks between different pages. We collect the WikiCS dataset and its raw text from~\cite{wikics} and OFA~\cite{OFA}. There are 11,701 nodes and 216,123 edges in the graph. The text feature format of nodes and edges is shown in the below:
\begin{tcolorbox}[boxsep=0mm,left=2.5mm,right=2.5mm, colback=themecolor]
\textbf{Node text:} Wikipedia entry. Entry name: $<$\textit{name}$>$. Entry content: $<$\textit{content}$>$.

\textbf{Edge text:} Page link between two Wikipedia entries.
\end{tcolorbox}
WikiCS dataset can be used for node classification. There are 10 categories for WikiCS, including Computational linguistics; Databases; Operating system; Computer architecture; Computer security; Internet protocols; Computer file systems; Distributed computing architecture; Web technology; Programming language topics. In the original dataset, there are 20 different splits, and we use the first set as our split, which results in 580/1,769/5,847 samples for the train/val/test sets, respectively. 

\subsection{Product-subset}
Product-subset is a co-purchase graph where nodes are product items from Amazon, and edges mean that two products are co-purchased. We collect Product-subset from TAPE~\citep{TAPE}. The original Product dataset is hosted by OGB~\citep{OGB}, which contains 2,449,029 nodes and 61,859,140 edges. In TAPE~\citep{TAPE}, the authors generate a subset from the original dataset. In Product-subset, there are 54,025 nodes and 144,638 edges.  The text feature format of nodes and edges is shown in the below:
\begin{tcolorbox}[boxsep=0mm,left=2.5mm,right=2.5mm, colback=themecolor]
\textbf{Node text:} Product from Amazon platform with title and content: $<$\textit{title}$>$$<$\textit{content}$>$.

\textbf{Edge text:} Connected two products are purchased together.
\end{tcolorbox}
Product-subset can be used for node classification, where the task is to classify the category of products. In the original source, the dataset contains 47 categories. However, in the subset created by TAPE, there is one class containing products from different categories with missing descriptions, and two classes without any node belonging to them. Therefore, we remove all samples from these three classes. After processing, there are a total of 14,695/1,567/36,982 samples for the train/val/test sets, respectively.

\subsection{FB15K237}
FB15K237 is a knowledge graph. The dataset contains 14,541 nodes and 310,116 relations. Nodes in the dataset are entities in the knowledge graph and edges represent the relation between two entities. We obtain FB15K237 from OFA~\citep{OFA}. The text feature format of nodes and edges is shown in the below:
\begin{tcolorbox}[boxsep=0mm,left=2.5mm,right=2.5mm, colback=themecolor]
\textbf{Node text:} Entity in the knowledge graph. Entity name: $<$\textit{name}$>$. Entity alternatives: $<$\textit{alternatives}$>$. Entity description: $<$\textit{descriptions}$>$.

\textbf{Edge text:} Relation from source entity to target entity: $<$\textit{relation}$>$.
\end{tcolorbox}

FB15K237 can be used for link prediction. It contains 237 different relation types. Following OFA~\citep{OFA}, we first convert it to an undirected graph and then split the dataset with a ratio of 0.85/0.05/0.1. This results in a total of 272,115/17,535/20,466 samples for train/val/test sets, respectively.

\subsection{WN18RR}
WN18RR is another knowledge graph extracted from WordNet. We obtained the dataset from OFA~\citep{OFA}. It contains 40,943 nodes and 93,003 relations where each node is an English word and each edge represents the relation between two words. The text feature format of nodes and edges is shown in the below:
\begin{tcolorbox}[boxsep=0mm,left=2.5mm,right=2.5mm, colback=themecolor]
\textbf{Node text:} English word and its description. Word name: $<$\textit{name}$>$. Word description: $<$\textit{description}$>$.

\textbf{Edge text:} Relation from source word to target word: $<$\textit{relation}$>$.
\end{tcolorbox}

WN18RR can be used for link prediction. It contains 11 different relation types. The split of WN18RR is obtained directly from OFA~\citep{OFA}. It contains 86,835/3,034/3,134 samples for train/val/test sets, respectively.

\subsection{MovieLens-1m}
MovieLens-1m is a recommendation graph containing nodes of users and movies. It is a bipartite graph where an edge indicates that the user rates the movie with a rating from 1 to 5. We obtained the MovieLens-1m from PyG~\citep{pyg}. It contains 9,923 nodes and 1,000,209 edges. The text feature format of nodes and edges is shown in the below:
\begin{tcolorbox}[boxsep=0mm,left=2.5mm,right=2.5mm, colback=themecolor]
\textbf{Movie node text:} Movie with title and genre. Title: $<$\textit{title}$>$. Genre: $<$\textit{genre}$>$.

\textbf{User node text:} User in the movie rating platform with the following information: gender: $<$\textit{gender}$>$, age: $<$\textit{age}$>$, occupation: $<$\textit{occupation}$>$.

\textbf{User-Movie edge text:} Source user rate the target movie with rating: $<$\textit{rate}$>$.

\textbf{Movie-User edge text:} Source movie rated by the target user with rating:  $<$\textit{rate}$>$.

\end{tcolorbox}

MovieLens-1m can be used for link regression or classification. The goal is to predict the rating score between users and movies. We randomly split all edges into train/val/test with a ratio of 0.85/0.05/0.1, which results in 850,177/50,011/100,021 samples for the train/val/test sets, respectively.

\subsection{Chembl molecular instruction}
Chembl is a collection of molecular datasets. It contains 13 different datasets ranging from molecule property classification to solubility regression. Specifically, the collection contains chemblpre; molproperties; PCBA; HIV; BBBP; BACE; toxcast; esol; freesolv; lipo; cyp450; tox21; and muv. The detailed statistics for each sub-dataset are included in Table~\ref{tab:dataset}. We obtain Chembl from GIMLET~\citep{gimlet}. For all datasets, each graph is a molecule, in which each node represents an atom, and each edge represents a chemical bond. The text feature format of nodes and edges is shown in the below:
\begin{tcolorbox}[boxsep=0mm,left=2.5mm,right=2.5mm, colback=themecolor]
\textbf{Node text:} Chemical atom with the following information: $<$\textit{atom information}$>$

\textbf{Edge text:} Chemical bond between two atoms with the following information: $<$\textit{bond information}$>$.
\end{tcolorbox}

Chembl datasets can be used for both classification and regression at the graph level. The split is obtained directly from GIMLET~\citep{gimlet}. Table~\ref{tab:chembl} shows the detailed split and task type.
\begin{table}[h]
\small
 \setlength{\tabcolsep}{8pt}
 \renewcommand{\arraystretch}{1}
 \setlength\tabcolsep{5pt}
    \caption{Chembl datasets task and task split. }
    \vspace{5pt}
    \label{tab:chembl}
    \centering
    \begin{tabular}{l|ccccc}
        \toprule
        Dataset & Train & Val & Test & Task type \\ \midrule
        Chemblpre & 341,952 & 0 & 0 & 1048-way binary classification\\
        molproperties &  363,336  & 0  & 0 & question answering \\
        PCBA & 349,854 & 43,650 & 43,588 & 128-way binary classification\\
        HIV & 32,901 & 4,113 & 4,113 & Binary classification\\
        BBBP & 1,631 & 204 & 204 & Binary classification\\
        BACE & 1,210 & 151 & 152 & Binary classification\\
        toxcast & 6,859 & 858 & 858 & 588-way binary classification\\
        esol & 902 & 113 & 113 & Regression\\
        freesolv & 513 & 64 & 65 & Regression\\
        lipo & 3,360 & 420 & 420 & Regression\\
        cyp450 & 13,516 & 1,690 & 1,690 &5-way binary classification\\
        tox21 & 6,264 & 783 & 784 & 12-way binary classification\\
        muv & 74,469 & 9,309 & 9,309 & 17-way binary classification \\
        \bottomrule
    \end{tabular}
\vspace{-18pt}
\end{table}

\subsection{ExplaGraphs}
ExplaGraphs is a graph question answering dataset on commonsense concepts. We obtain ExplaGraphs from G-retriever~\citep{gretriever}. Each graph in ExplaGraphs contains commonsense concepts connected by its relation. The text feature format of nodes and edges is shown in the below:
\begin{tcolorbox}[boxsep=0mm,left=2.5mm,right=2.5mm, colback=themecolor]
\textbf{Node text:} Common sense concept: $<$\textit{concept}$>$

\textbf{Edge text:} Common sense relation: $<$\textit{relation}$>$.
\end{tcolorbox}

ExplaGraphs can be used for question-answering on graphs. The task is to answer whether the given two arguments are counter or support. We obtain the split directly from G-retriever~\citep{gretriever}. It contains 1,659/553/554 graph samples for the train/val/test sets.

\subsection{SceneGraphs}
SceneGraphs is a graph question-answering dataset on scene graphs. We obtain SceneGraphs from G-retriever~\citep{gretriever}. Each graph in SceneGraphs contains objects connected by the relationship of two objects. The text of nodes and edges is in the following format:
\begin{tcolorbox}[boxsep=0mm,left=2.5mm,right=2.5mm, colback=themecolor]
\textbf{Node text:} Object in an image. Name: $<$\textit{name}$>$; attribute: $<$\textit{attribute}$>$; (x,y,w,h): $<$\textit{coordinates}$>$.

\textbf{Edge text:} Relation between two objects: $<$\textit{relation}$>$.
\end{tcolorbox}

SceneGraphs can be used to answer questions on graphs. The questions focus on asking about the properties of objects or the relation between two objects in the graph. We obtain the split directly from G-retriever~\citep{gretriever}. It contains 59,978/19,997/20,025 graph samples for the train/val/test sets.

\subsection{MAG240M-subset}
MAG240m-subset is a citation graph between academic papers. The original MAG240M is generated by OGB-LSC~\citep{OGB-lsc}. It contains 121,751,666 papers, 122,383,112 authors, 25,721 institutions, and more than 1.3 billion edges. However, the original dataset is too large and the raw text for author nodes and institution nodes is hard to collect. Thus, we sample a subset from the original one. Specifically, we only include paper nodes and citation edges. Further, we exclude all nodes not in the train/val/test sets provided by the original source. For edges, we exclude all edges whose two ends are not in the train/val/test sets. After the processing, there are a total of 5,875,010 nodes and 26,434,726 edges. The text feature format of nodes and edges is shown in the below:
\begin{tcolorbox}[boxsep=0mm,left=2.5mm,right=2.5mm, colback=themecolor]
\textbf{Node text:} Academic paper with title and abstract: $<$\textit{title}$>$$<$\textit{abstract}$>$.

\textbf{Edge text:} Connected two papers have a citation relationship.
\end{tcolorbox}

MAG240M-subset can be used for node classification. The goal is to classify the categories of papers. There are 153 categories. We obtain the split directly from OGB-LSC~\citep{OGB-lsc}. There are a total of 900,722/126,675/132,585 samples in the train/val/test sets. 

\subsection{Ultrachat200k}
Ultrachat200k is a graph question-answering dataset. we obtain the original Ultrachat200k from~\citep{ultrachat}. It is a multi-round conversation dataset used for tuning large language models. In TAGLAS, we only use the \textit{train-sft} subset and convert it to a graph question-answering dataset by creating chain graphs on it. In particular, suppose one sample in the original Ultrachat200k has $k$ rounds of conversation. We create $k-1$ different graphs. The first graph contains two nodes and one edge, where one node contains the question in the first round, and the other node contains the answer of that round. The edge links from the question to the answer, where the question and answer of the second round will be the target question and answer. Similarly, the second graph contains four nodes and three edges with a similar format. After the processing, there are a total of 449,929 graphs. The text feature format of nodes and edges is shown in the below:
\begin{tcolorbox}[boxsep=0mm,left=2.5mm,right=2.5mm, colback=themecolor]
\textbf{Question node text:} $<$\textit{question}$>$.

\textbf{Answer node text:} $<$\textit{answer}$>$.

\textbf{Q-A edge text:} This edge represents the target sentence answer to the instruction in the source sentence.

\textbf{A-Q edge text:} This edge represents the target sentence as an instruction followed by the source answer.
\end{tcolorbox}

Ultrachat200k dataset can be used for graph question-answering. We split the dataset into train/val/test sets with 400,000/20,000/29,929 samples in each set.

\subsection{Wikikg90m}

Wikikg90m is an encyclopedic knowledge graph dataset extracted from wikidata knowledge base. We obtain the original Wikikg90m from OGB-LSC~\citep{OGB-lsc}. It contains 91,230,610 entities, 1,387 relations, and 601,062,811 edges. The text of nodes and edges is in the following format:

\begin{tcolorbox}[boxsep=0mm,left=2.5mm,right=2.5mm, colback=themecolor]
\textbf{Node text:} Entity in the knowledge graph. Entity name: $<$\textit{name}$>$. Entity description: $<$\textit{descriptions}$>$.

\textbf{Edge text:} Relation from source entity to target entity. Relation title: $<$\textit{title}$>$. Relation description: $<$\textit{descriptions}$>$
\end{tcolorbox}

Wikikg90m can be used for link prediction. The goal is to classify the relation type between two entities. We obtain the split directly from OGB-LSC. There are a total of 601,062,811/15,000 triplets in the train/val sets. OGB conserves the test set answer for evaluation purposes, therefore, we do not include it here.

\subsection{Dataset loading}
Loading of a dataset in TAGLAS is fairly simple. We annotate each dataset with a unique key and users can load a specific dataset through the key. We also provide API for users to load multiple datasets at the same time. A simple demonstration is shown below. 
\begin{mintedbox}{python}
from TAGLAS import get_dataset, get_datasets
# Load Arxiv dataset.
dataset = get_dataset("arxiv")
# Load multiple datasets.
dataset_list = get_datasets(["arxiv", "pcba"])
\end{mintedbox}
\section{Task construction}
In this section, we describe the construction of the task in TAGLAS. Basically, the tasks are divided into three different levels: node level, link level, and graph level. Typically, each dataset is associated with one or multiple task levels. Further, each task level can be divided into three different task types: default, subgraph, and question answering, to serve different model training and evaluation scenarios. Note that, for each task type, we further provide two versions: non-text version and text version. For the non-text version, the task will retrieve the node and edge feature from its original source. For datasets that don't have original features, the identical feature will be generated. The text version loads the raw text feature for nodes and edges. 

\subsection{Node-level tasks}
\textbf{Default format}: The default format implements the most common task format for node-level semi-supervised prediction~\citep{GCN, gat, SGC}. Specifically, the default format will directly return the whole graph. There will be an indicator to indicate whether a particular node is in the split and the label of it. Models or algorithms for this task type are supposed to be run directly on this whole graph and return the predictions of all nodes in the split simultaneously. 

\textbf{Subgraph format}: In the real world, many graphs are super large (perhaps millions of nodes and billions of edges), which are hard to fit into the memory. Therefore, the subgraph task format is designed. Instead of returning the whole graph, in the subgraph format, we sample a small subgraph rooted at the node we want to predict. Consequently, the task becomes to make a prediction for the target node based on the sampled subgraph. Recently, it has been shown that the subgraph format has many advantages like improved expressivity~\citep{ngnn, zeng2021decoupling, zhang2021labeling} and unified task representation~\citep{allinone, OFA, liu2023graphprompt}. In TAGLAS, we adopt the random sampling. Users can specify the number of hops and the maximum number of nodes per hop. Then, iterative sampling is done on each hop. Given sampled nodes in the $i$-th hop, to sample the $i+1$-th hop, the algorithm first extracts all nodes connected to nodes in the $i$-th hop. If the number of nodes exceeds the pre-defined maximum threshold $k$, we will randomly select $k$ nodes at this hop. After the sampling of nodes, the algorithm will extract all edges between the sampled nodes to construct the final subgraph. 

\subsection{Link-level tasks}
\textbf{Default format}: Similar to the default format in the node level, the default format in the link level directly returns the whole graph and the labels of the corresponding split for the models. However, due to the lack of link-level expressivity of this format~\citep{zhang2021labeling}, this task type is mainly used for running the baseline. 

\textbf{Subgraph format}: The subgraph format in the link level will sample a subgraph for each link. In TAGLAS, we first independently sample a subgraph for both the two end nodes using the same procedure described in the node-level tasks. Next, the two subgraphs are merged to form the final subgraph for the link. 

\subsection{Graph-level tasks}
For graph-level tasks, the prediction is made upon the whole graph and no subgraph sampling is needed. Therefore, we only provide the default format for the graph level, where each graph is a data sample (like a molecule). 

\subsection{Question answering tasks}
Finally, TAGLAS also supports question-answering tasks. In question-answering tasks, instead of tensor/index-based labels, each sample may related to a question-answer pair represented by natural language. In TAGLAS, some datasets are created for the purpose of question-answering, like SceneGraphs or Ultrachat200k. They are naturally supported by question-answering tasks. Further, for all tasks described above, the original labels are provided in the tensor format. However, we can still convert the task into free-form question-answering format. To accommodate this, TAGLAS provides a question-answering version for each type of task so that each sample will associate it with a question and an answer. For example, for node classification in citation networks, we may ask ``What is the category of the paper?'', and the answer is the textual description of the label of the target node. Note that for node/edge level question answering tasks, we use subgraph-based task format by default. 

\subsection{Task loading}
We provide a simple API for loading different tasks. Similarly, users can specify the dataset, task type, and other parameters (like split or subgraph sampling parameters) in the API.

\begin{mintedbox}{python}
from TAGLAS import get_task
# Load default node-level task on Cora.
task = get_task("cora_node", "default")
# Load text version subgraph edge-level task on PubMed dataset and valid split.
task = get_task("pubmed_link", "subgraph_text", split="val")
\end{mintedbox}

\section{Other Features}
\subsection{Text to embedding conversion}
For all text-related tasks, we provide an API to convert the raw text features to sentence embedding through LLMs. This could benefit the evaluation of GNN-based models. We implement several commonly used LLMs, including BERT~\citep{BERT}, sentence transformer~\citep{SentenceEncoder}, E5~\citep{E5}, Llama2-7b, and Llama2-13b~\citep{llama2}. For all language models, we employ mean pooling to compute sentence embedding from token embeddings. Users can also define their own sentence embedding models. The computed embeddings will be saved for future use. Below is a sample code:
\begin{mintedbox}{python}
from TAGLAS import get_task
from TAGLAS.tasks.text_encoder import SentenceEncoder
# Load sentence transformer.
encoder_name = "ST"
encoder = SentenceEncoder(encoder_name)
arxiv_task = get_task("arxiv", "subgraph_text", split="test")
# Convert raw text in Arxiv dataset to sentence embedding.
arxiv_task.convert_text_to_embedding(encoder_name, encoder)
\end{mintedbox}

\subsection{Advanced subgraph sampling}

In default, for all tasks that are in the subgraph format, the subgraph is randomly sampled at each hop. However, this approach can disrupt the original distribution of the graph. To address this, we introduce advanced subgraph sampling methods, including support for personalized PageRank. Specifically, we first compute importance scores for each node using the personalized PageRank algorithm. During the sampling stage, the probability of selecting a node at each hop is weighted according to its importance score relative to the other nodes in that hop. Below is a sample code:

\begin{mintedbox}{python}
from TAGLAS import get_task
# use personalized page rank for subgraph sampling.
arxiv_task = get_task("arxiv", "subgraph_text", split="test", use_ppr_sampling=True)
\end{mintedbox}

\subsection{Evaluation}
To facilitate the evaluation of a model on the collated datasets, TAGLAS provides various evaluation metrics built upon the torchmetrics~\citep{torchmetrics}. Particularly, we provide an evaluation API for evaluating both models with tensor output and models with text output. For each dataset, we select the most commonly used metric for evaluation. Similarly, all APIs can be obtained easily through the key of dataset and the type of task:
\begin{mintedbox}{python}
from TAGLAS import get_evaluator
# Get default evaluator for cora_node task. metric_name is a string indicates the name of metric.
metric_name, evaluator = get_evaluator("cora_node", "subgraph_text")
# Get QA evaluator for arxiv
metric_name, evaluator = get_evaluator("arxiv", "QA")
\end{mintedbox}
\section{Conclusion}
In this report, we present TAGLAS, the first dataset collection focused on text-attributed graphs. In TAGLAS, we collect more than 23 TAG datasets from various domains and task levels. TAGLAS further implements task construction, text-to-embedding conversion, evaluation, and more API for fast and easy-to-use training and evaluation on both GNN-based and LLM-based models. We hope the presence of TAGLAS could contribute to the research of both the graph and NLP communities. The project is still in development. Please expect more datasets and features in the future.

\medskip
\newpage
\bibliography{references}
\bibliographystyle{unsrtnat}

\end{document}